\documentclass[11pt]{article}

\usepackage[preprint]{acl}

\usepackage{times}
\usepackage{latexsym}
\usepackage{amsmath}
\usepackage{amsfonts}

\usepackage[T1]{fontenc}
\usepackage[utf8]{inputenc} 

\usepackage{microtype}
\usepackage{inconsolata}

\usepackage{graphicx}

\title{When Dimensionality Hurts: The Role of LLM Embedding Compression for Noisy Regression Tasks}

\author{Felix Drinkall*, Janet B. Pierrehumbert*\ddag, Stefan Zohren* \\
    *Department of Engineering Science, University of Oxford \\
    \ddag Faculty of Linguistics, University of Oxford \\
    \texttt{felix.drinkall@eng.ox.ac.uk}}

\begin{document}
\maketitle
\begin{abstract}
Large language models (LLMs) have shown remarkable success in language modelling due to scaling laws found in model size and the hidden dimension of the model's text representation. Yet, we demonstrate that compressed representations of text can yield better performance in LLM-based regression tasks. In this paper, we compare the relative performance of embedding compression in three different signal-to-noise contexts: financial return prediction, writing quality assessment and review scoring. Our results show that compressing embeddings, in a minimally supervised manner using an autoencoder's hidden representation, can mitigate overfitting and improve performance on noisy tasks, such as financial return prediction; but that compression reduces performance on tasks that have high causal dependencies between the input and target data. Our results suggest that the success of interpretable compressed representations such as sentiment may be due to a regularising effect.
\end{abstract}


\section{Introduction}

Modern machine learning research increasingly relies on LLMs to handle complex real-world tasks \cite{ 10.5555/3692070.3693283, rahimikia2024r, huang-etal-2024-enhancing}. Recent progress in LLM performance has largely come from scaling models' parameters, training dataset size and the expressivity of the LLM via the model's hidden dimension \cite{kaplan2020scalinglawsneurallanguage}. In recent years the hidden dimension has scaled from a standard representation size of 768 dimensions \cite{devlin-etal-2019-bert} to up to 16384 \cite{grattafiori2024llama3herdmodels}, and possibly higher in some large closed-source models. While it is not clear whether the relationship between model scaling and linguistic performance will hold indefinitely \cite{10.5555/3666122.3668712}, it is generally accepted that modelling language requires a high-dimensional representation space \cite{grattafiori2024llama3herdmodels}. 
This has meant that LLMs have very strong formal linguistic competence \cite{MAHOWALD2024517}. However, some machine learning tasks, like stock returns prediction tasks, have inherently low signal-to-noise relationships between the input and output \cite{sawhney-etal-2020-deep}, which we will refer to as ``noisy'' tasks in this paper. In the case of predicting stock returns from news, noise arises not only from uninformative articles or weak causality between the article and the stock price but can also come from delayed reactions, market efficiency, and unpredictable macroeconomic influences \cite{antweiler2004all, MantillaGarcia2017}. In such noisy research areas, the link between dimensionality and performance is unclear, and feature selection or compression can act as a regularising component \cite{TIAN2022146}. When input dimensionality is too large, models risk overfitting by memorising noise rather than learning meaningful patterns. On the other hand, a low-dimensional embedding might underfit by losing critical high-order interactions. This paper explores the relationship between text-embedding dimensionality and downstream performance in tasks where the signal is noisy, focusing on a stock returns prediction task.

Not all tasks are well-suited to purely generative LLMs; many tasks benefit more from supervised machine learning \cite{tang2024understandingllmembeddingsregression}, where labelled data guide classification or regression outcomes by identifying robust dependencies between input text and target outputs \cite{johan-berggren-etal-2019-regression}.  Generative models often require extensive computational resources and large datasets \cite{10.5555/3600270.3602446}, creating obstacles under computational or data constraints. Problems also emerge when using the output from generative models in larger architectures that fuse textual data with other modalities, such as numerical or structured information \cite{drinkall2025forecastingcreditratingscase}, since generative models can produce unpredictable outputs \cite{10.1145/3491102.3517582} and are relatively weak at complex numerical reasoning \cite{liu-etal-2024-llms}. As such, it is often preferable to use the embedding representations from LLMs as input features for a conventional neural network in regression-based tasks as opposed to passing all of the numerical and textual data into a prompt.

There have been papers that have investigated how well text embeddings perform in regression tasks \cite{tang2024understandingllmembeddingsregression}, but none have investigated the degree to which the noise of a task affects the optimal dimensionality of the text representation. This is despite the widespread adoption and success of interpretable, compressed representations of text in financial return prediction tasks like sentiment \cite{math10132156}, emotion \cite{TILLY2021114760}, and topics \cite{drinkall-etal-2022-forecasting, Garc_a_M_ndez_2023}. While many papers have shown that these compressed representations can perform better than raw text embeddings, this paper investigates the degree to which this is due to regularisation rather than the true value of the features. We consider the internal representation from an autoencoder, and show that interpretable features such as sentiment or emotion do not deliver an improvement beyond text representation compression to a more optimal dimensionality.

This paper has the following contributions:

\begin{enumerate}
    \item Providing some evidence for a link between the optimal representational dimensionality of text in a regression task and the signal-to-noise ratio of the dataset.
    \item Demonstrating that gains attributed to interpretable features (e.g., emotion, sentiment) in financial returns tasks may primarily stem from representational compression, rather than from inherently superior feature sets.
    \item Identifying the optimal compression dimensionality of text representation in a financial returns task.
    \item A financial news stock returns dataset\footnote{\url{https://github.com/FelixDrinkall/financial-news-dataset/}} released under an academic license which can be used for benchmarking purposes.
\end{enumerate}

\section{Datasets}
\label{sec:datasets}

To explore the relationship between the signal-to-noise ratio and the optimal dimensionality of the text representation in a regression task, we compare three domains of conceivable regression tasks. Stock market return prediction using news articles is a notoriously noisy domain \cite{black1986noise} since a significant proportion of the news articles are likely to not contain any useful information \cite{antweiler2004all}. We contrast this noisy domain to customer review and essay marking datasets, which both have a very strong connection between the regression input and the target value.

\subsection{Stock Returns Dataset}
\label{subsec:stock_dataset}

We combine two data sources to form the dataset: CRSP daily price data\footnote{\url{https://tinyurl.com/mrxsfdhu}} and news articles. For the 50 most traded U.S. stocks, we use the closing bid/ask average as the daily price. Given the previous day's closing price $p_{t-1}$ and the next day's closing price $p_{t+1}$, the daily return is defined as:

\[
r_{t} = \frac{p_{t+1}-p_{t-1}}{p_{t-1}}.
\]

This return $r_{t}$ serves as the regression model’s target. We use the next and previous day's data as opposed to the current day's price data so that we can be sure that the publication of the article intersected the two prices, and thus avoid including samples where the article was published after the market's closing time. The underlying assumption is that the news content is either causing or reflecting the observed price change. 

We source news articles via CommonCrawl News\footnote{\url{https://commoncrawl.org/}} \cite{Hamborg2017}, scraping articles from Yahoo Finance. Using a pre-trained named entity recognition BERT model \cite{tjong-kim-sang-de-meulder-2003-introduction, devlin-etal-2019-bert}, we extract all mentioned organizations, then filter them through a dictionary of company abbreviations to identify target companies. We then apply another filter to make sure that only one of the target companies is mentioned in each sample to reduce the noise slightly. The size and split for the datasets are reported in Appendix \ref{app:stock_ret_data}.

\subsection{High Signal Datasets}

To compare the degree to which noise affects the optimal dimensionality of a task we selected a dataset with a high causal dependency between the input data and target information. Written product reviews are directly linked to the score assigned to the review, therefore we use the following datasets: from Yelp \cite{NIPS2015_250cf8b5}, App Reviews \footnote{\url{https://tinyurl.com/427hnu6c}} and Amazon Reviews \cite{McAuley2013}. Writing quality is a more ambiguous task but there is still a direct link between the input text and the target which also makes it a good candidate for comparison against more noisy tasks -- as such we use the ELL English Grading dataset \cite{Franklin2022}. Details of each dataset are in Appendix \ref{app:dataset_details}.

\section{Methodology}



Given the success of generative LLMs, much of the recent research on downstream tasks has focused on how to use LLMs in a prompting setting \cite{chang2024efficientpromptingmethodslarge}. However, there are some domains where encoder-based LLMs are better suited: embedding tasks have been dominated by LLMs pre-trained with bi-directional attention \cite{NEURIPS2020_c3a690be} or uni-directional attention followed by bi-directional fine-tuning \cite{lee2024nvembedimprovedtechniquestraining, llm2vec}. Likewise, recent work shows that generative models perform worse on word meaning comprehension than encoder-based LLMs \cite{qorib-etal-2024-decoder}. As such, we encode the textual information using \textit{all-mpnet-base-v2} \cite{NEURIPS2020_c3a690be, reimers-gurevych-2019-sentence}, an encoder-based model fine-tuned using a contrastive objective function on a series of sentence similarity tasks. The model is widely used and competitive on the MTEB benchmark \cite{muennighoff-etal-2023-mteb}.

\subsection{Input Processing and Chunking}

Each textual input \( x \) is tokenized into a sequence of tokens \((t_1, t_2, \ldots, t_{L})\). To handle variable-length inputs, we split the token sequence into \(M\) chunks, each of length at most \(C\), where \(C\) is the maximum context window of the model.
If the final chunk is shorter than \(C\), it is padded; similarly, sequences with fewer than \(M\) chunks are padded, ensuring each batch element has uniform shape \([M, C]\). After chunking, we pass the sequences through a pre-trained language model to produce token-level embeddings, and then mean-pool across the final layer token representations of all of the chunks to produce \(\mathbf{v}_i \in \mathbb{R}^{d_{\text{LLM}}}\). Where $d_{\text{LLM}}$ is the size of the LLM's hidden dimension. The target features are standardized to enable easier interpretation.

\subsection{Dimensionality Reduction}

To obtain a lower-dimensional representation, we train an autoencoder consisting of an encoder \( E: \mathbb{R}^{d_{\text{LLM}}} \to \mathbb{R}^{d_{z}} \) and a decoder \( D: \mathbb{R}^{d_{z}} \to \mathbb{R}^{d_{\text{LLM}}} \), where $d_{\text{z}}$ is the size of the autoencoder's hidden dimension:

\[
\mathbf{z}_i = E(\mathbf{v}_i), \quad \hat{\mathbf{v}}_i = D(\mathbf{z}_i).
\]

The autoencoder is trained to minimize:

\[
\mathcal{L}_{\text{AE}} = \frac{1}{N}\sum_{i=1}^{N} \|\mathbf{v}_i - \hat{\mathbf{v}}_i\|^2.
\]

The training runs for 100 epochs with early stopping if validation loss does not improve for 5 epochs. After training, we use \(E\) to compress all embeddings \(\mathbf{v}_i\) into \(\mathbf{z}_i \in \mathbb{R}^{d_{z}}\) and vary $d_{\text{z}}$ to identify the optimal compression ratio.

The dimensionality reduction methodology in this section is benchmarked against more interpretable techniques that are widely used in stock returns tasks such as emotion and sentiment scores. We take the softmax outputs from fine-tuned DistilRoBERTa models that have been fine-tuned on sentiment and emotion classification tasks \footnote{Emotion: \url{https://tinyurl.com/mrxxnuft}} \footnote{Sentiment: \url{https://tinyurl.com/3upb7zyw}} and pass the class probabilities into the regression model. This enables us to compare their relative performance against the autoencoder latent features to infer the degree to which their strong performance is due to regularisation or valuable feature selection.

\subsection{Regression Model}

For the regression model, we use a random forest model as it is robust, lightweight and widely used \cite{Breiman2001, Roy01122012}, which simplifies the experimental setup so that focus can be directed to the compression methodology. For the same reason we use the default parameters and do not vary them between runs. We also tried a two-layer MLP layer in line with \citet{tang2024understandingllmembeddingsregression} (Appendix \ref{app:MLP}), but the inadequate overall performance and high variance in prediction errors meant that no statistically significant conclusions could be drawn. 

\section{Results}
\label{sec:results}

We report the results using a Huber loss function \cite{huber1964robust} since it is robust to outliers while maintaining sensitivity to small errors, whereas Mean Squared Error (MSE) is often dominated by the performance of outliers. The Huber threshold between L1 and L2 loss that we use is 1. The significance level of $d_{\text{z}}=x$ is determined using a T-Test \cite{student1908probable} between the Huber error distributions of the best performing latent dimension $d^*_{\text{z}}$ and $d_{\text{z}}=x$.

\begin{figure}[t]
    \centering
    \includegraphics[width=\linewidth]{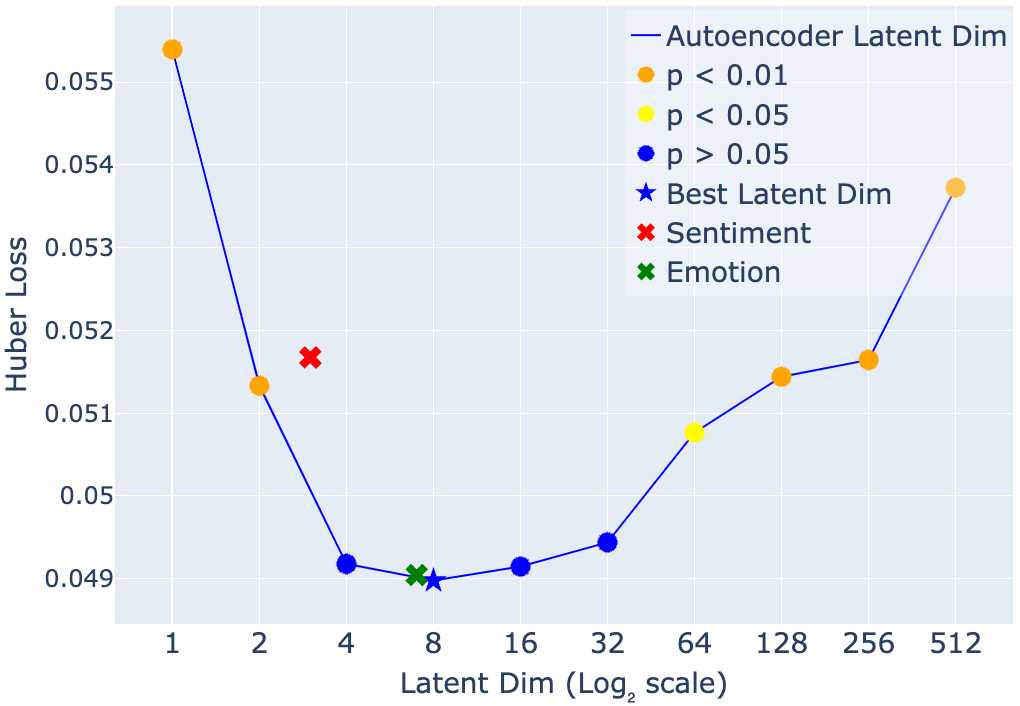}
    \caption{Huber Loss on the noisy financial returns task for different autoencoder latent dimensions $d_{\text{z}}$. The performance of \textcolor{red}{sentiment} and \textcolor{OliveGreen}{emotion} representations also appear inline with their respective dimensionality. The significance of each result compared to the best latent dimension is displayed using \textcolor{Blue}{blue} (p > .05), \textcolor{YellowOrange}{orange} (p < .01) and \textcolor{Yellow}{yellow} (p < .05) colours.}
    \label{fig:fin_returns_task}
\end{figure}

By varying the hidden dimension of the autoencoder $d_{\text{z}}$ and then passing the input into our regression model, Figure \ref{fig:fin_returns_task} shows that the optimal dimensionality on the financial returns task is 8, but that there is not a statistically significant difference between a $d_{\text{z}}=8$ and $d_{\text{z}} \in \{4, 16, 32\}$. There is also a significant difference between $d_{\text{z}}=8$ and $d_{\text{z}} \in \{1, 2, 128, 256, 512\}$.  The result shows us that the optimal dimensionality is significantly less than $d_{\text{LLM}}$, showing that for this noisy task some dimensionality reduction is necessary. 

Figure \ref{fig:fin_returns_task} also reports the performance of compressing the text representation into a class probability vector of different semantic and emotion categories. Both representations do not exceed the expected performance of the autoencoder features at their respective dimensions $d_{\text{z}}$. Despite the reported success of emotion and sentiment features in similar tasks \cite{TILLY2021114760, math10132156} the findings of this work suggest that most of this performance improvement can be explained due to the regularising effect of dimensionality reduction rather than the inherent value of the features.

\subsection{Impact of Noise}
\label{sec:res_noise}

\begin{figure}
    \centering
    \includegraphics[width=1\linewidth]{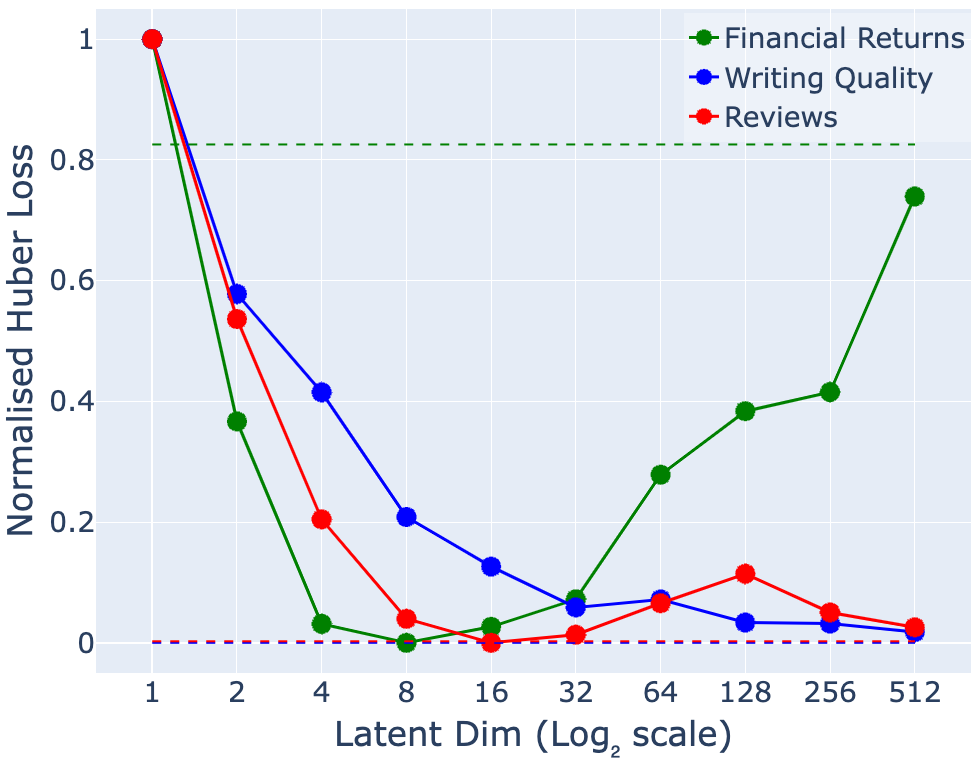}
    \caption{Huber loss averaged over \textcolor{red}{Review}, \textcolor{blue}{English Writing}, and \textcolor{OliveGreen}{Financial Returns} tasks - granular performance in Appendix \ref{app:dataset_results}. To display the results on one axis, the y-axis represents the error rate of each individual task as a percentage of the maximum and minimum error on that task. The performance without any compression is marked with the dashed line.}
    \label{fig:noise-comp}
\end{figure}

To compare the degree to which noise affects the optimal dimensionality of a task we test our methodology on tasks from different domains. Figure \ref{fig:noise-comp} shows that for the financial returns task, there is a convex relationship between performance and dimensionality, whereas the relationship approximates a negative exponential in tasks with a strong signal; the performance does not deteriorate for high dimensional inputs. The large difference in error distributions between the different tasks suggests that input dimensionality is a key parameter for regression-based tasks. 

Also of note is that in all domains the performance reaches 10\% of the minimum loss at a much lower dimension than $d_{\text{LLM}}$. We will call the dimension at which this performance is achieved the "intrinsic dimension", which for the Review and the English Writing tasks is 8 and 32 respectively. This suggests that the pertinent signals for regression tasks in general can be compressed to a lower dimension space and still achieve strong performance. For any architectures which have poor time complexity as a function of input length, this is an important finding.

\section{Conclusion}

Our results suggest that the optimal dimensionality is dependent on the signal-to-noise ratio, exposing the necessity of feature compression in high-noise environments. The implication of the result is that researchers should consider the noise of a task when making decisions about the dimensionality of text. In particular, the results highlight the importance of dimensionality reduction in financial returns prediction tasks, with an optimal autoencoder latent dimension of $d_{\text{z}}=8$. The lack of statistical significance for $d_{\text{z}} \in \{4, 16, 32\}$ suggests some flexibility in choosing the dimensionality, while extreme values lead to significant performance deterioration. Additionally, the findings indicate that sentiment and emotion-based representations do not provide inherent advantages over learned latent features, implying that their previous success in similar tasks may be attributed to regularisation effects rather than intrinsic informativeness.

\section{Future Work}

Future research could explore adaptive dimensionality compression methods that dynamically adjust based on the signal-to-noise ratio; however, to do this, a measure of signal-to-noise is required before processing the input features. Researchers could also use this framework to assess the relative performance of new text encoding methodologies in regression tasks to make sure that the value does not just come from model regularisation. It would be desirable to further explore the unsupervised autoencoder features and the more classic emotion features in more downstream settings.

\section{Limitations}

Although our findings demonstrate the importance of reducing dimensionality in high-noise tasks, several limitations should be noted. Firstly, the paper does not explore other ways of compressing the text representations. PCA, t-SNE, UMAP \cite{vandermaaten08a, 2018arXivUMAP} are all options for dimensionality compression. This current version of the paper does not provide comparisons with these techniques, but we will provide the comparison in future versions of this paper. Secondly, while the experiments include multiple domains for high-signal datasets, the study focuses on a specific financial returns dataset with a single definition of “noisy” data; future work could explore a broader range of domains. Thirdly, while we aimed to keep the modelling process simple to not distract from the main thrust of the paper - data compression methods - the findings are only reported for one model.

\section*{Acknowledgements}

The first author was funded by the Economic and Social Research Council of the UK via the Grand Union DTP. This work was supported in part by a grant from the Engineering and Physical Sciences Research Council (EP/T023333/1). We are also grateful to the Oxford-Man Institute of Quantitative Finance and the Oxford e-Research Centre for their support. 

\bibliography{acl_latex}

\appendix

\section{Dataset Details}
\label{app:dataset_details}

\subsection{Stock Returns Dataset}
\label{app:stock_ret_data}

As outlined in the main section of the paper, this dataset was curated for this paper. Most of the dataset details are outlined in Section \ref{subsec:stock_dataset}, but this section contains the details that are missing. The test set is the whole of 2023, which contains 17,810 articles, and the training and validation set are defined using a temporal split which takes the last 10\% of data between 2017 and 2022. The resultant training and validation sets contain 30,115 and 3,346 samples respectively.

\subsection{Yelp Reviews Dataset}
\label{app:yelp_data}

The Yelp dataset \cite{NIPS2015_250cf8b5} consists of 700k Yelp reviews with a star rating between 1 and 5. There are 650k training samples and 50k testing samples, and the split is taken from the Huggingface dataset \textit{Yelp/yelp\_review\_full}.

\subsection{App Reviews Dataset}
\label{app:app_data}

\begin{figure*}[t]
    \centering
    \includegraphics[width=\linewidth]{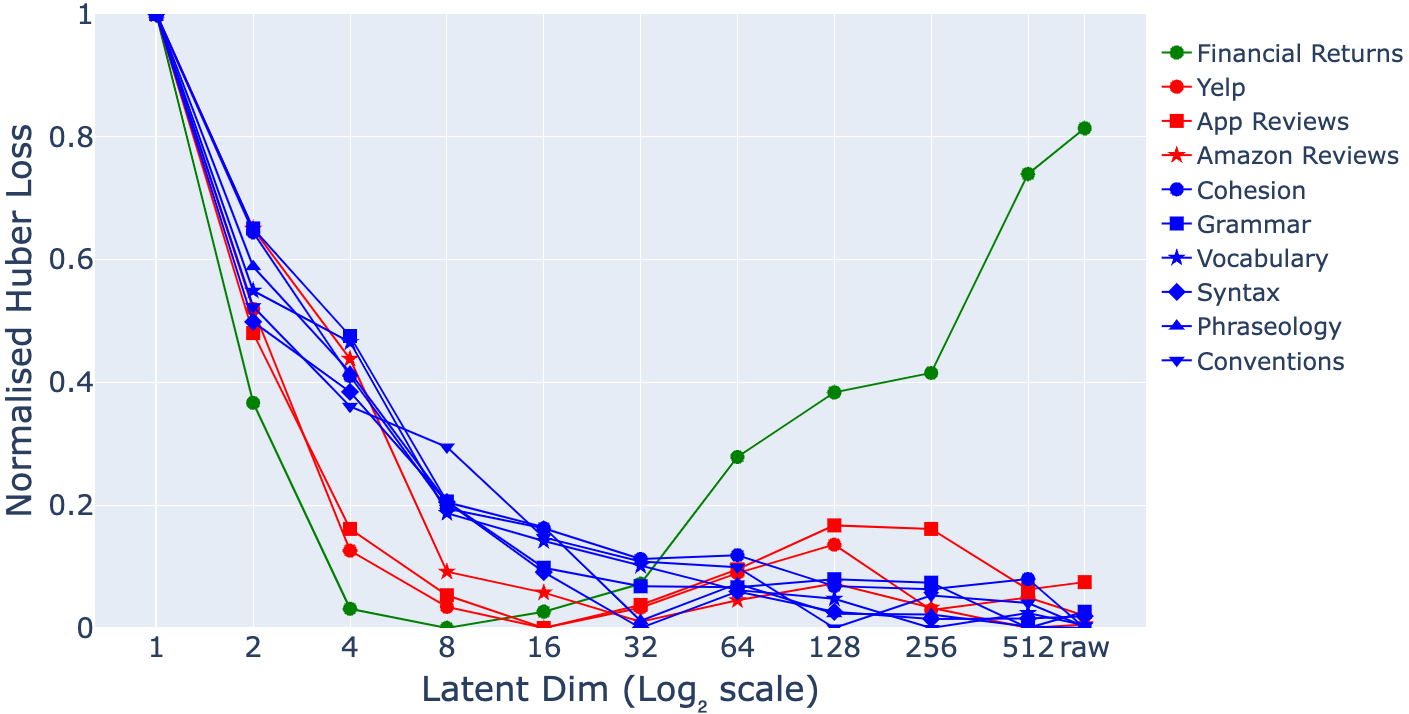}
    \caption{The normalised Huber loss of each dataset that makes up the result in Figure \ref{fig:noise-comp}. "raw" appears in the same location as 768 in the plot since this is the dimension of the non-compressed embedding.}
    \label{fig:ind_daatset_results}
\end{figure*}

This App Review dataset contains reviews of 395 Android applications, covering 629 versions. It provides the review and the star rating between 1 and 5, and user-reported issues in English. The dataset consists of 288k samples and we perform a random train-val-test split of 85.5:4.5:10.

\subsection{Amazon Reviews Dataset}
\label{app:amazon_data}

The Amazon Reviews dataset \cite{McAuley2013} consists of 568k fine food reviews collected from Amazon over a period of more than 10 years, up to October 2012. Each review includes a product ID, user ID, profile name, rating (1–5), helpfulness votes, timestamp, summary, and full text. The data came from the Huggingface dataset \textit{jhan21/amazon-food-reviews-dataset}, which did not contain any predefined train-test splits so we performed a random train-test split of 85.5:4.5:10.

\subsection{Writing Quality Dataset}
\label{app:writing_data}

The writing quality dataset \cite{Franklin2022} comes from a Kaggle competition set up by Vanderbilt University. The competition aimed to improve automated feedback tools for English Language Learners (ELLs) by developing language proficiency models using real student essays. The dataset assesses English text over six criteria: cohesion, syntax, vocabulary, phraseology, grammar and conventions. We report the results of cohesion, vocabulary, and grammar. The dataset consists of 3.91k samples and we perform a random train-val-test split of 85.5:4.5:10.

\section{Individual Dataset Results}
\label{app:dataset_results}

Figure \ref{fig:ind_daatset_results} shows the performance of each dataset that makes up the averaged result in Figure \ref{fig:noise-comp}.

\section{MLP}
\label{app:MLP}

The configuration for the unsuccessful MLP is outlined in this section. The model was not able to learn the financial returns task for any dimensional input. We believe that this negative result will aid other researchers in this area. The compressed embeddings \(\mathbf{z}_i\) serve as inputs to an MLP with hidden dimension \( d_{\text{mlp}} \):

\[
\mathbf{h}^{(1)} = \mathcal{D}_p \bigl(\sigma(W^{(1)}\mathbf{z}_i + \mathbf{b}^{(1)})\bigr),
\]
\[
\mathbf{h}^{(2)} = \mathcal{D}_p \bigl(\sigma(W^{(2)}\mathbf{h}^{(1)} + \mathbf{b}^{(2)})\bigr),
\]
\[
\hat{y}_i = W^{(3)}\mathbf{h}^{(2)} + \mathbf{b}^{(3)},
\]

where \(\mathbf{z}_i \in \mathbb{R}^{d_{z}}\), \( W^{(1)} \in \mathbb{R}^{d_{\text{mlp}} \times d_{z}} \), and \( W^{(2)}, W^{(3)} \) similarly match the required dimensions. Dropout \(\mathcal{D}_p(\cdot)\) is applied with probability \(p\), and \(\sigma(\cdot)\) is the ReLU activation. We optimize the Huber loss with \(\delta=1.0\):

\[
\mathcal{L}_{\text{reg}} = \frac{1}{N}\sum_{i=1}^{N} \text{HuberLoss}(y_i, \hat{y}_i, \delta).
\]

Most targets \(y_i\) are close to zero, so we apply target standardisation and stop training when validation loss does not improve for 5 epochs, restoring the best model state.

\section{Autoencoder Visualisation}
\label{app:reconstruction_visual}

\begin{figure}[b!]
    \centering
    \includegraphics[width=\linewidth]{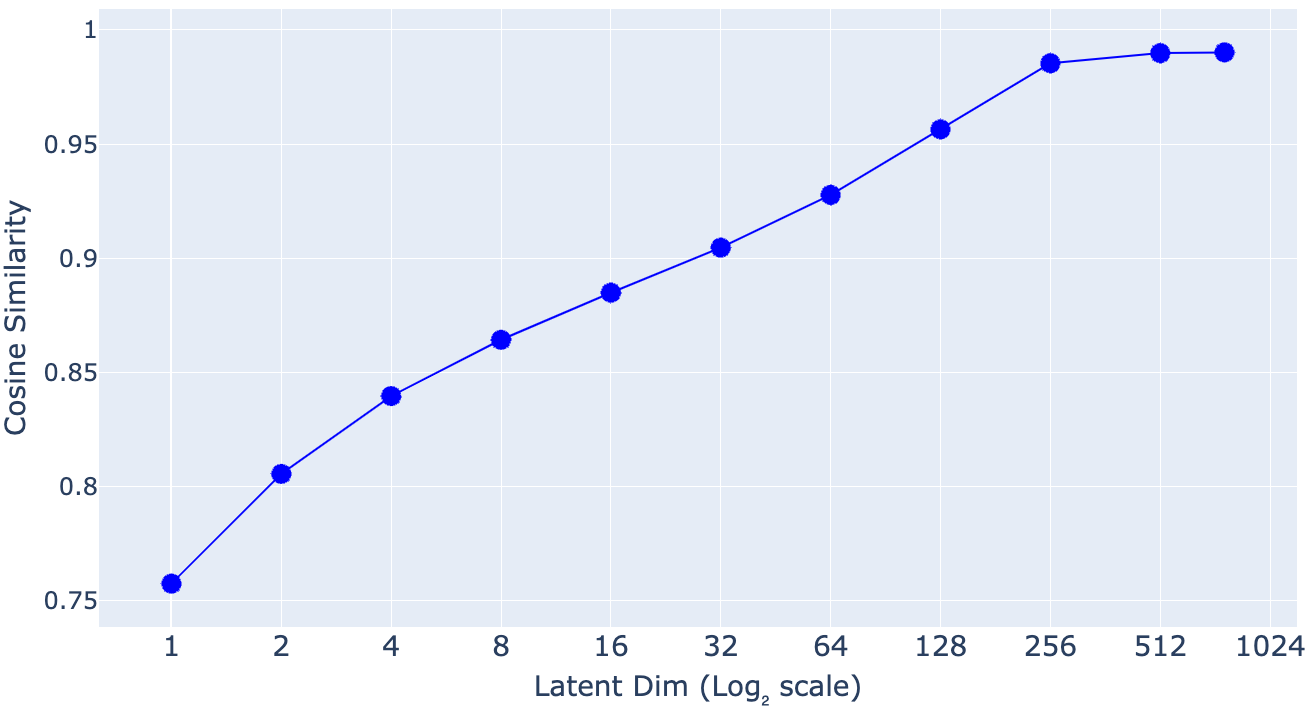}
    \caption{Cosine similarity between $\mathbf{v}_i$ and $\hat{\mathbf{v}}_i$ on the financial returns dataset.}
    \label{fig:cosine_sim}
\end{figure}

Downstream performance on regression tasks provides insight into the quality of the autoencoder's compression. However, Figure \ref{fig:cosine_sim} offers a more direct comparison between the autoencoder's input and output embeddings, $\mathbf{v}_i$ and $\hat{\mathbf{v}}_i$, respectively. The figure displays the cosine similarity between the raw ($\mathbf{v}_i$) and reconstructed ($\hat{\mathbf{v}}_i$) embeddings for different hidden dimensions. The graph provides us with a further understanding of the reconstructive process; it seems that a $d_z=256$ is the point at which performance reaches an asymptote. It also implies that there is some semantic loss at the optimal dimensions in Figure \ref{fig:noise-comp}. 

\end{document}